%% file: main.tex
\def\year{2022}\relax
\documentclass[letterpaper]{article} 
\usepackage{aaai22}  
\usepackage{times}  
\usepackage{helvet}  
\usepackage{courier}  
\usepackage[hyphens]{url}  
\usepackage{graphicx} 
\usepackage{natbib}  
\usepackage{caption} 
\DeclareCaptionStyle{ruled}{labelfont=normalfont,labelsep=colon,strut=off} 
\usepackage{algorithm}
\usepackage{algorithmic}

\usepackage{newfloat}
\usepackage{listings}
\floatstyle{ruled}
\newfloat{listing}{tb}{lst}{}
\floatname{listing}{Listing}
\title{Open Vocabulary Electroencephalography-To-Text Decoding and Zero-shot Sentiment Classification}

\author{
    Zhenhailong Wang,\textsuperscript{\rm 1}
    Heng Ji\textsuperscript{\rm 1}
}
\affiliations{
    
    \textsuperscript{\rm 1}University of Illinois at Urbana-Champaign\\
    
    \{wangz3, hengji\}@illinois.edu
}

\input{init}

\begin{document}
\maketitle

\input{Abstract}
\input{0intro}
\input{1task_formulation}
\input{2method}
\input{3experiment}
\input{4related_works}
\input{5conclusion}
\section*{Acknowledgement}
This research is based upon work supported in part by U.S. DARPA GAILA Program No. HR00111990058. The views and conclusions contained herein are those of the authors and should not be interpreted as necessarily representing the official policies, either expressed or implied, of DARPA, or the U.S. Government. The U.S. Government is authorized to reproduce and distribute reprints for governmental purposes notwithstanding any copyright annotation therein.

\bibliography{main}

\clearpage
\newpage


\appendix
\input{Appendix_section}

\end{document}

%% file: init.tex
\usepackage{graphicx}               
\usepackage{tabularx}               
\newcolumntype{C}{>{\centering\arraybackslash}X}
\usepackage{multirow}               
\usepackage{diagbox}                
\usepackage{hhline}                 
\usepackage{amsmath}                
\usepackage{amssymb}                
\usepackage{mathtools}              
\usepackage{enumitem}               
\newcommand{\bs}{\boldsymbol}

\usepackage{subfigure}
\usepackage{booktabs}
\usepackage{extarrows}
\usepackage{makecell}

\usepackage{footmisc} 

\usepackage{xcolor}
\usepackage{xparse}
\usepackage{hyperref} 

\newcommand{\Zhenhailong}[1]{\textcolor{blue}{\textit{Zhenhailong:} \textit{#1}}}
\renewcommand{\Zhenhailong}[1]{} 

\NewDocumentCommand{\TODO}{ mO{} }{\textcolor{green}{\textsuperscript{\textit{TODO}}\textsf{\textbf{\small[#1]}}}}

%% file: Abstract.tex
\begin{abstract}
State-of-the-art brain-to-text systems have achieved great success in decoding language directly from brain signals using neural networks. However, current approaches are limited to small closed vocabularies which are far from enough for natural communication. Additionally, most of the high-performing approaches require data from invasive devices (e.g., ECoG). In this paper, we extend the problem to open vocabulary Electroencephalography(EEG)-To-Text Sequence-To-Sequence decoding and zero-shot sentence sentiment classification on natural reading tasks. We hypothesize that the human brain functions as a special text encoder and propose a novel framework leveraging pre-trained language models (e.g., BART). Our model achieves a 40.1\% BLEU-1 score on EEG-To-Text decoding and a 55.6\% F1 score on zero-shot EEG-based ternary sentiment classification, which significantly outperforms supervised baselines.
Furthermore, we show that our proposed model can handle data from various subjects and sources, showing great potential for a high-performance open vocabulary brain-to-text system once sufficient data is available.\footnote{The code is made publicly available for research purpose at \url{https://github.com/MikeWangWZHL/EEG-To-Text}. \textbf{[Edit] There is an unintentional bug in the original evaluation code which makes the results in the paper not as expected. Please refer to the README in the Github Repo for more details before using the codebase.}} 
\end{abstract}

%% file: 0intro.tex
\section{Introduction}
\label{sec:introduction}
Understanding and decoding the human brain has always been a fascinating topic for centuries. In recent years, Brain-Computer-Interface (BCI) based on motor imagery has gained great success in helping paralytic people restoring motor functionalities such as reaching and grasping~\cite{hochberg2012reach, aflalo2015decoding, bouton2016restoring}. Decoding natural language from brain signals, on the other hand, remains a major challenge. We point out that previous approaches on brain-to-text and brain-to-speech decoding~\cite{herff2015brain, anumanchipalli2019speech, makin2020machine, sun2019towards, panachakel2021decoding,  nieto2021thinking,  doi:10.1056/NEJMoa2027540} still have limitations in terms of vocabulary size, device, and articulation dependency, etc. (Section~\ref{sec:related_works}). 
Previous work mainly focuses on achieving high accuracy, thus decodes sentences and words in small closed vocabularies. Moreover, current systems lack the ability to decode semantically close words that do not exist in the training set.
In this paper, we extend the problem from closed vocabulary to open vocabulary Electroencephalography(EEG)-To-Text Sequence-To-Sequence decoding as well as zero-shot sentiment classification on natural reading tasks. We enlarge the vocabulary size for more than 100 times from several hundred~\cite{makin2020machine, sun2019towards} to 50,265\footnote{We use pretrained BART vocabulary:  \url{https://huggingface.co/transformers/model_doc/bart.html?highlight=bart\#transformers.BartConfig} \label{fn:vocab}}. We utilize data~\cite{hollenstein2018zuco,hollenstein2020zuco} from various subjects and sources recorded by non-invasive devices.

Previous work on brain-to-speech~\cite{herff2015brain, anumanchipalli2019speech, makin2020machine, doi:10.1056/NEJMoa2027540} decoding has successfully captured low-level auditory features from the movement of our vocal tract to reconstruct words.
Instead of only capturing articulation features, another line of work demonstrated that the human brain encodes language into higher dimensional semantic representations~\cite{gauthier2018does, correia2014brain}. Interestingly, we have seen similar behavior in large-scale pretrained language models, such as BERT~\cite{devlin2018bert}, BART~\cite{lewis2019bart}, GPT2~\cite{radford2019language} and GPT3~\cite{brown2020language}, which encode words into contextualized semantic embeddings~\cite{ethayarajh2019contextual}. Recent findings on multimodal neurons~\cite{goh2021multimodal} in CLIP~\cite{radford2021learning} revealed another level of resemblance between artificial neurons and human brain neurons in the sense that they all respond to highly abstract concepts. The major contributions of the aforementioned large-scale pretrained language models are their transfer learning abilities. By fine-tuning them on specific downstream tasks, we observe a substantial improvement in various NLP tasks, including sequence classification, text generation, etc. Although covariate shift has been generally observed in brain signal data due to intra- and inter-subject variability~\cite{lund2005motion, saha2020intra}, previous work demonstrated promising transfer learning ability in brain signal decoding using deep learning models~\cite{roy2020deep, zhang2020application, makin2020machine}.
Furthermore, various studies~\cite{muttenthaler2020human, hollenstein2021decoding, hollenstein2019cognival, hale2018finding, schwartz2019understanding} have experimented with connecting brain signal decoding to NLP models, by either using brain signals as an additional modality for improving performance on NLP tasks or using NLP models to understand how the human brain encodes language. 

In this paper, we extend previous work to a new level by using pretrained language models for open vocabulary EEG-to-text decoding. The motivation of using pretrained language models is that contextualized representation from pretrained language models carries important linguistic information, including syntactic features, semantic features, and long-distant dependencies~\cite{tenney2018what,jawahar2019does}. This existing knowledge obtained from observing large text corpora is particularly useful for our open vocabulary decoding task with scarce data. Based on the assumption that the human brain functions as a special text encoder, we leverage pretrained language models via jointly fine-tuning with additional projection layers.
We then further evaluate its power on a novel zero-shot sentence-level sentiment classification task. 
Detailed definition of the new tasks can be found in Section~\ref{sec:task_formulation}.

In terms of the choice of device, although invasive devices like Electrocorticography (ECoG) generally result in better performance~\cite{burle2015spatial}, we focus on using non-invasive EEG data for a few reasons. Compared to other non-invasive devices like functional magnetic resonance imaging (fMRI), EEG has relatively high temporal resolution with an affordable cost~\cite{zanzotto2010comparing, hecht2015techniques, yi2013multi}. Analogous to training a language model, the sheer amount of data is essential for learning representation of brain signals~\cite{brown2020language}. Compared to data from invasive-devices, EEG data is easier to acquire and more publicly available. However, unlike the abundance of text and image data, we have far from enough brain-text paired data to train a high-performance open vocabulary brain-to-text sequence-to-sequence model. Nevertheless, we have observed a trend in growing availability for open-source EEG-integrated devices\footnote{\url{https://openbci.com/}}, which can be a huge potential data source. 
And we suggest that, shortly, non-invasive BCI will have a larger and wider impact on everyone's life as a new type of human-machine interaction tool in various areas, including Virtual Reality and Augmented Reality~\cite{putze2020brain}. 

To summarize, the main contributions of this paper are as follows.
\begin{itemize}
    \item We introduce two new tasks: Open vocabulary EEG-To-Text decoding and EEG-based sentence sentiment classification. To the best of our knowledge, this is the first work using open vocabulary setting on brain-to-text decoding.
    \item We are the first to use pretrained language models for EEG-To-Text decoding. We further propose a novel zero-shot pipeline for EEG-based sentiment classification. 
    \item We show that our proposed framework can leverage data from various subjects and sources, which demonstrates great potential for high-performance open vocabulary EEG-To-Text system. 
\end{itemize}

%% file: 1task_formulation.tex
\section{Task Definition}
\label{sec:task_formulation}
\input{symbol}
\begin{figure}[t]
    \centering
    \includegraphics[width=\linewidth]{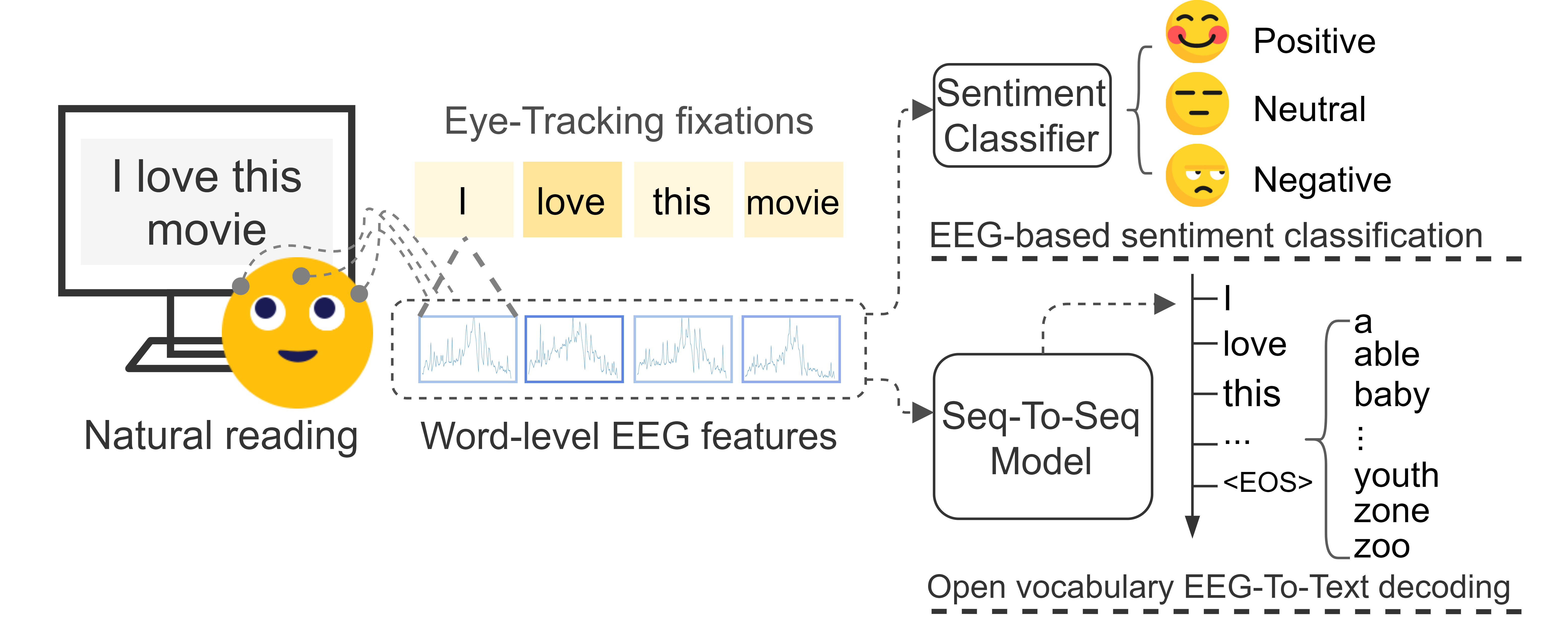}
    \caption{An overview of proposed tasks. Subjects are asked to read text on a screen at their own speed. Simultaneous eye-tracking and EEG data are recorded. The darker the background color, the more fixations are on the word. Word-level EEG features can be extracted by synchronizing with eye-tracking fixations. EEG feature sequences then serve as inputs for sequence-to-sequence decoding or sentiment classification. In this paper, we use ZuCo datasets for experiments, please refer to Section~\ref{subsec:experiment_dataset} for more details.
    }
    \label{fig:task}
\end{figure}

\subsubsection{Open vocabulary EEG-To-Text decoding} 
Given a sequence of word-level EEG features $\mathcal{E}$, the task is to decode the corresponding text tokens from an open vocabulary $\mathcal{V}$ in a Sequence-To-Sequence framework. In this paper, we use EEG-Text pairs $\langle \mathcal{E}, \mathcal{S} \rangle$ recorded in natural reading tasks, e.g., ZuCo dataset~\cite{hollenstein2018zuco, hollenstein2020zuco}.
During the training phase, such EEG-Text pairs can come from various subjects and various categories of reading materials. During the test phase, the text sentences $\mathcal{S}$ in $\langle \mathcal{E}, \mathcal{S} \rangle$ are totally unseen.

\subsubsection{EEG-based sentence sentiment classification}
Given aforementioned word-level EEG feature sequences $\mathcal{E}$, the task is to predict the sentiment label $c \in \mathcal{C}$ of the corresponding text sentence $\mathcal{S}$. Instead of including text input~\cite{hollenstein2021decoding, kumar2019fusion} along with brain signal features, we use EEG features as our only input. We further introduce \textbf{zero-shot EEG-based sentence sentiment classification}, in which we do not require any EEG-Sentiment pairs $\langle \mathcal{E}, c \rangle$. Instead, we use only EEG-Text pairs $\langle \mathcal{E}, \mathcal{S} \rangle$ and external Text-Sentiment pairs $\langle \mathcal{S}_{ext}, c \rangle$. 
Thus, we can leverage EEG-Text pairs without sentiment labels from various sources, as well as existing text-sentiment datasets, such as Yelp\footnote{\url{https://www.yelp.com/dataset}} and Stanford Sentiment Treebank\cite{socher2013recursive}. An overview of the proposed tasks can be found in Figure~\ref{fig:task}.

%% file: symbol.tex
\begin{table}[t]
\small
    \centering
    \begin{tabular}{c|m{17em}}
    \toprule 
       \textbf{Symbol}  & \textbf{Meaning} \\
       \midrule
 $v \in \mathcal{V}$         & English token in an open vocabulary\\
 $\textbf{e} \in \mathcal{E}$         & EEG feature vector in an EEG sequence \\
 $c \in \mathcal{C}$         & Sentiment label in ternary sentiment classes\\
 $\langle \mathcal{E}, \mathcal{S} \rangle$         & Word-level EEG feature sequence and text sentence pair\\
 $\langle \mathcal{E}, c \rangle$         & Word-level EEG feature sequence and sentiment label pair\\
 $\langle \hat{\textbf{e}}, c \rangle$         & Aggregated sentence-level EEG and sentiment label pair\\
 $\langle \mathcal{S}, c \rangle$         & Text sentence and sentiment label pair\\
    \bottomrule
    \end{tabular}
    \caption{List of symbols}
    \label{tab:symbols}
\end{table}

%% file: 2method.tex
\section{Method}
\label{sec:method}
\begin{figure*}[t]
    \centering
    \includegraphics[width=\linewidth]{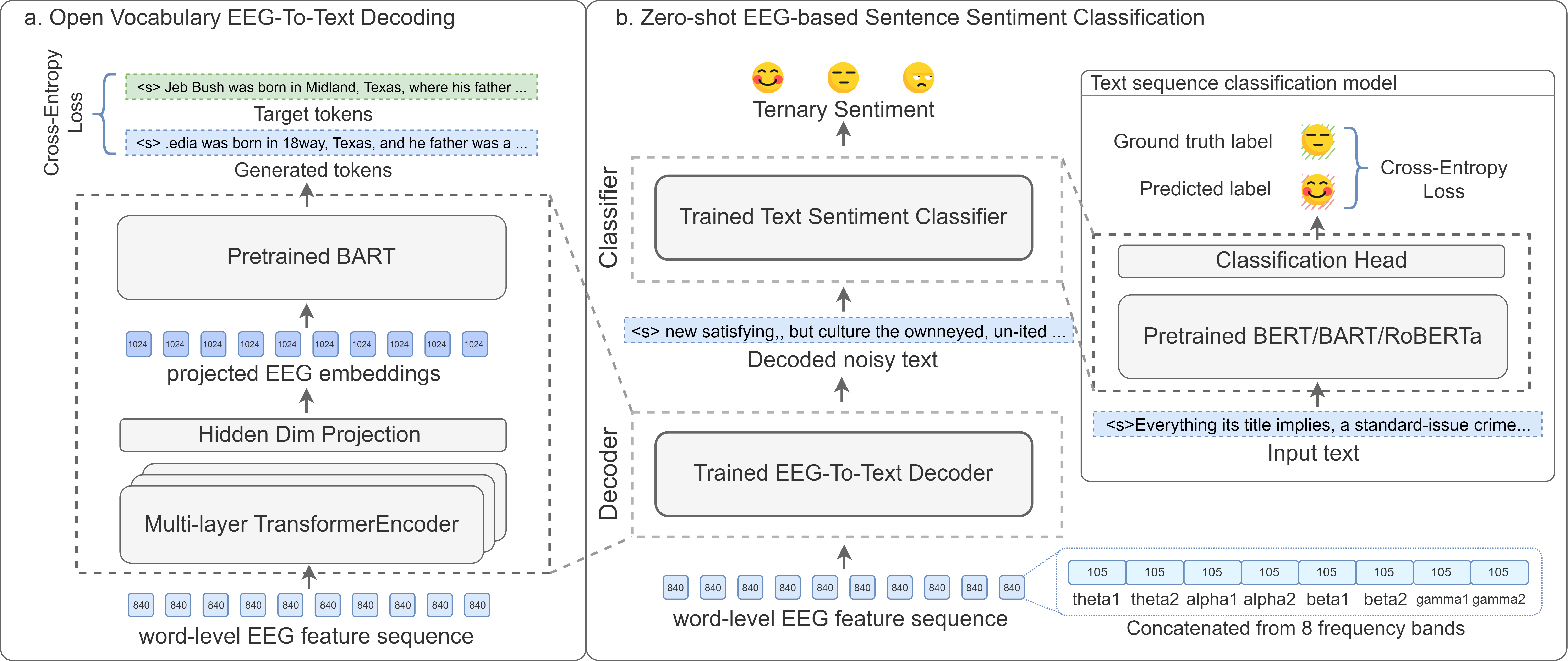}
    \caption{(a) Our proposed framework for EEG-To-Text decoding. (b) Our proposed pipeline for zero-shot EEG-based sentence sentiment classification. EEG-To-Text model trained on EEG-Sentence pairs in (a) is plugged in as Decoder in (b). Sequence classification model trained on external Text-Sentiment pairs is plugged in as Classifier in (b). 
    }
    \label{fig:framework}
\end{figure*}

\subsection{EEG-To-Text Decoding}
\label{sec:method_generation}
Similar to \cite{makin2020machine}, we formulate the EEG-To-Text Decoding task in a neural machine translation setting~\cite{sutskever2014sequence, bahdanau2014neural}. We try to maximize the probability of the decoded sentence: 
\begin{align}
    p(\mathcal{S}|\mathcal{E}) = \prod_{t = 1}^{T}p(s_t \in \mathcal{V}|\mathcal{E},s_{<t})
\end{align}
where $T$ is the length of the target text sequence.
The main challenge in our setting is that our vocabulary size $|\mathcal{V}|$ ($\sim50000$) is significantly larger than previous sequence-to-sequence studies ($\sim250$)~\cite{makin2020machine}. In addition, we use more noisy non-invasive EEG data. To address these challenges, we propose a novel framework leveraging pretrained language models. Inspired by the machine translation application using pretrained BART as described in ~\cite{lewis2019bart}, we treat each EEG feature sequence as an encoded sentence by the human brain. We then train an additional encoder to map the embedding from the human brain to the embedding from the pretrained BART. The high-level idea is that we assume the human brain to be a special kind of encoder, as mentioned in Section~\ref{sec:introduction}, which functions similar to a language model that encodes a sequence of English tokens into contextualized embeddings. 
One major difference from a traditional machine translation task is that the tokens in the input sequence here are not drawn from a fixed vocabulary, e.g. French words, but rather a continuous feature space. Inspired by Speech Recognition~\cite{hinton2012deep}, where the acoustic input is also represented as continuous vectors, we use these EEG feature vectors directly as initial word embeddings to feed into the model. 

As shown in Figure~\ref{fig:framework}.a, the model contains two major components, a randomly initialized additional encoder, and a pretrained encoder-decoder BART. Given a sequence of EEG features $\bs{h}_e$, we first feed them into a Multi-layer Transformer Encoder (MTE) and then a single layer feed-forward network to get the mapped embedding $\bs{h}_m$ of the sequence. Then the mapped embedding is fed into pretrained BART encoder and decoder. Finally, the last hidden states from the BART decoder are fed into a fully connected layer to generate English tokens $s_t$ from pretrained BART vocabulary $\mathcal{V}$.
\begin{align}
    \bs{h}_m &= \text{ReLU}\left ((\text{MTE}(\bs{h}_e))^T\bs{W}_e \right)\\
    p(s_t \in \mathcal{V}) &= \text{Softmax}\left ( \text{BART}(\bs{h}_m)^T\bs{W}_d \right)\\
    \mathcal{L}_{\text{rec}} &= -\sum_t^{T} \log p(s_t \in \mathcal{V}) \label{eq:loss}
\end{align}
where the MTE has 6 layers and 8 attention heads, $\bs{W}_e$ represents the weights from the fully connected projection layer, $\bs{W}_d$ represents the weights from the language modeling head, which is a single fully connected layer. All pooling methods use the last hidden states in both MTE and BART decoders. During training, the objective is to minimize the text reconstruction cross-entropy loss as shown in equation~\ref{eq:loss}. 

\subsection{Zero-shot Sentiment Classification Pipeline} \label{sec:method_zeroshot}
We further propose a novel pipeline leveraging the aforementioned Sequence-To-Sequence EEG-To-Text model to do zero-shot EEG-based sentence-levelsentiment analysis. As shown in Figure~\ref{fig:framework}.b, the proposed pipeline consists of two modules, a decoder, and a classifier. The idea is to first use an EEG-To-Text model to convert EEG features into noisy text, and then use a text-based classifier to predict the sentiment label. That makes this pipeline zero-shot in the sense that both the decoder and the classifier are trained individually, on EEG-Text pairs $\langle \mathcal{E}, \mathcal{S} \rangle$ and external Text-Sentiment pairs $\langle \mathcal{S}, c \rangle$  respectively, and thus no EEG-Sentiment pairs $\langle \mathcal{E}, c \rangle$ are required. Moreover, the modularized architecture makes it easy to upgrade with improved decoders and classifiers. In this paper, we use a BART-based~\cite{lewis2019bart} decoder as mentioned in the previous section. We also experiment with different choices of classifiers based on pretrained BERT~\cite{devlin2018bert}, BART~\cite{lewis2019bart}, and RoBERTa~\cite{liu2019roberta}. 

The idea of first decoding EEG features into text comes from observations on some preliminary experiments. We find that baselines trained explicitly on EEG-Sentiment pairs $\langle \mathcal{E}, c \rangle$ do not work well on decoding sentiment directly from EEG features (Section~\ref{sec:experiment_sentiment}). We hypothesize that, compared to previous emotion analysis studies~\cite{koelstra2011deap, liu2020eeg} where EEG signals are recorded from image/video stimulus, it is more difficult to directly decompose sentiment related features from text semantic features in our natural reading EEG data. However, in our zero-shot pipeline, decoding EEG sequence into text serves as an information filter, which enables us to further leverage the power of fine-tuned text-based sentiment classifiers. 

%% file: 3experiment.tex
\section{Experiment}
\label{sec:experiment}
\subsection{Dataset}
\label{subsec:experiment_dataset}
\input{dataset}
We use ZuCo~\cite{hollenstein2018zuco, hollenstein2020zuco} datasets, which contain simultaneous EEG and Eye-tracking data recorded from natural reading tasks. The reading tasks include Normal Reading (NR) and Task-Specific Reading (TSR). Reading material is collected from movie reviews~\cite{socher2013recursive} and Wikipedia  articles. Normal reading with movie reviews has ground-truth ternary sentiment labeling: positive, neutral, or negative. In this paper, we use concatenated word-level EEG feature sequences aggregated by gaze duration (GD). For more details on EEG input data please refer to the Appendix. We further clean up the dataset by omitting sentences that contain NaN values. And then we split each reading task's data into \textit{train}, \textit{development}, \textit{test} (80$\%$,10$\%$,10$\%$) by unique sentences, that is, the sentences in \textit{test} set are totally unseen. The final statistics of the dataset can be found in Table~\ref{tab:dataset}.

\subsection{EEG-To-Text Decoding Evaluation}
\label{sec:experiment_generation}
We train our EEG-To-Text model on EEG-Text sequence pairs. Analogous to typical NLP fine-tuning tasks, we hypothesize that the model should be able to benefit from expansion of the training corpora. So we gradually increase the training dataset size by adding data from various reading tasks. We also include data from different subjects to test the model's robustness against inter-subject variability. We report the BLEU scores and ROUGE-1 scores in Table~\ref{tab:generation_scores}.
\input{generation_scores}
\subsubsection{Results}
\label{sec:experiment_metric_results}
The results show that, by increasing the scale of the training set, the model achieves significantly better performance. More importantly, we find that the model can handle data from various subjects and materials. By comparing models trained on ``\textit{SR v1.0 (half${_{1st}}$)}" with ``\textit{SR v1.0 (half${_{1st}}$)+NR v1.0 (half${_{1st}}$)}" and ``\textit{SR v1.0}", we find that the model achieves more improvements when adding data from diverse source (BLEU-1: from 33.5$\%$ to 37.3$\%$) than from uniform source (BLEU-1: from 33.5$\%$ to 34.5$\%$). And we observe similar behavior even when adding data from different subjects as shown by comparing ``\textit{SR v1.0 (half${_{1st}}$)}" with ``\textit{SR v1.0 (half${_{1st}}$)+NR v1.0 (half${_{2nd}}$)}" (BLEU-1: from 33.5$\%$ to 37.4$\%$). Furthermore, our best-performed model is trained and tested on ``\textit{SR v1.0$+$NR v1.0$+$NR v2.0}" data, which is from 30 different subjects reading movie reviews and Wikipedia articles. 

We also observe that the choice of the reading task is important. Comparing models trained on ``\textit{SR v1.0$+$NR v1.0}" and on ``\textit{SR v1.0$+$NR v1.0$+$TSR v1.0}", we can see that adding Task-Specific Reading (TSR) data does not result in visible improvements. The reason is that in the TSR setting, subjects are asked to read text prompt with a preliminary question such as ``Does this sentence contain the \textit{Nationality} relation?". We reason that such a setting can make subjects focus on specific parts of a sentence, which results in highly different distributions in EEG features compared with data from Normal Reading tasks (SR, NR). In addition, we showcase the power of using a pretrained language model by reporting scores from the same model trained from scratch (\textit{w/o pretrained weights}). 

\input{generation_results}
\subsubsection{Discussion}
\label{sec:experiment_generation_results}
By taking a closer look at the decoding results (Table~\ref{tab:generation_results}), we find that the model can sometimes correctly capture named entities, such as ``\textit{George W. Bush}" in (1) and ``\textit{Puerto Rico}" in (2), which does not even exist in the training set. 
In the cases that the model fails to decode the correct entity mentions, e.g., ``\textit{San Juan[LOCATION]}" vs ``\textit{New Francisco[LOCATION]}" in (2), ``\textit{Adolf Otto Reinhold Windaus[PERSON]}" vs ``\textit{Hitler[PERSON]}" in (5), the entity types are correctly captured. 

To systematically evaluate this interesting behavior, we perform named entity recognition\footnote{We use off-the-shelf named entity recognizer from Spacy: \url{https://spacy.io/usage/linguistic-features\#named-entities}} on both ground truth and model output. We calculate a \textit{longest common subsequence (LCS)} based matching score and a \textit{multiset cosine similarity (Multiset)} score on the extracted sequences of entity types, as shown in equation~\ref{eq:lcs} and~\ref{eq:multiset}. 

\begin{align}
    Sim_{LCS} &= \frac{\text{LCS}(X,Y)}{\text{max}(|X|,|Y|)}\label{eq:lcs}
    \\
    Sim_{Multiset} &= \frac{\sum x_i y_i}{\sqrt{\sum x_i^2} \sqrt{\sum y_i^2}}, i \in \mathcal{I}\label{eq:multiset}
\end{align}
where $X, Y$ are sequences of named entity types, e.g. Geo-political entities (GPE), PERSON, from ground truth and model output respectively. $LCS(X,Y)$ returns the length of the longest common subsequence of $X, Y$. $x_i, y_i$ are the number of instances on type $i$. $\mathcal{I}$ is a union of the sets of extracted entity types from ground truth and model output. The results can be found in Table~\ref{tab:entity_matching}. 
\input{entity_matching}

Apart from that, we find that the model can generate semantically close words or synonyms even though they do not exactly match the ground truth, as shown in \textit{Italic} words in Table~\ref{tab:generation_results}. For example, in (5), although ``\textit{monsterous}" does not appear in the training set and is actually a typo, the model is able to generate ``\textit{bad}" which semantically resembles ``\textit{monstrous}". 
Because the available training data is far from enough for fine-tuning a large language model like BART, the model still struggles to correctly reconstruct most of the fine-grained entities, as shown in the \underline{underlined} words in Table~\ref{tab:generation_results}. And the decoded sentences still suffer from grammatical errors. 

\subsection{EEG-based Sentiment Classification Evaluation}
\label{sec:experiment_sentiment}
We first implement a few baselines that are explicitly trained on EEG-Sentiment pairs $\langle \mathcal{E}, c \rangle$. And then we evaluate the baselines and our zero-shot pipeline proposed in Section~\ref{sec:method_zeroshot} on the same test set from reading task SR v1.0.
\subsubsection{Baselines}
\label{sec:experiment_sentiment_baseline}
\begin{itemize}
    \item \textbf{MLP}\hspace{1em}A Multi-Layer Perceptron baseline with three fully connected layers, ReLU activation, and one dropout layer. MLP is trained on aggregated sentence-level EEG-Sentiment pairs $\langle \hat{\textbf{e}}, c \rangle$, where $\hat{\textbf{e}}$ is an average over all word-level EEG features in a sequence.
    \item \textbf{Bi-LSTM}\hspace{1em}A Bi-directional-LSTM baseline with four stacked LSTM layers and a single-layer classification head.
    \item \textbf{BERT}\hspace{1em}BERT-based baselines with additional multi-layer transformer encoders (MTE). We also experiment with training from scratch using randomly initialized weights and fine-tuning from pretrained weights. Both LSTM and BERT baselines are trained on EEG-Sentiment sequence pairs $\langle \mathcal{E}, c \rangle$ from reading task SR v1.0.
    \item \textbf{Text Baselines}\hspace{1em}For more comprehensive comparison, we also report scores from text-based sentiment classification models based on pretrained BERT/BART/RoBERTa. We train and test these baselines on Text-Sentiment pairs $\langle \mathcal{S}, c \rangle$ from reading task SR v1.0.
    
\end{itemize}

\input{sentiment_scores}
\subsubsection{Pipeline}
In our zero-shot pipeline, we experiment with various combinations of decoders and classifiers. 
For the decoder, we use the BART-based EEG-To-Text decoding model from Section~\ref{sec:experiment_generation} trained on SR v1.0 EEG-Text pairs $\langle \mathcal{E}, \mathcal{S}\rangle$. We demonstrate the importance of the additional Multi-layer Transformer Encoder (MTE) component by removing it from the model. We also experiment with the decoder using a randomly initialized BART model.  
For the classifier, we experiment with sequence classification models based on BERT, BART and RoBERTa, which have identical architectures as the text baselines mentioned in Section~\ref{sec:experiment_sentiment_baseline}. Here, instead of being trained on SR v1.0 data, the classifiers are trained on a subset of Text-Sentiment pairs $\langle \mathcal{S}_{ext}, c \rangle$ from Stanford Sentiment Treebank~\cite{socher2013recursive}. We include sentences with a sentiment score in the range of very negative ($[0,0.2]$), neutral ($(0.4,0.6]$) or very negative ($(0.8,1.0]$), and assign a ternary sentiment label $c \in \{0,1,2\}$ to them respectively. And then we exclude those sentences that are already in the ZuCo SR v1.0 dataset to make sure that the sentences $\mathcal{S}_{ext}$ in $\langle \mathcal{S}_{ext}, c \rangle$ do not overlap with the sentences $\mathcal{S}$ in $\langle \mathcal{E}, \mathcal{S} \rangle$, which are used to train the Decoder. This setting guarantees that we do not require any EEG-Sentiment pairs $\langle \mathcal{E}, c \rangle$. The results are presented in Table~\ref{tab:sentiment_scores}.
\subsubsection{Results}
One major observation from EEG-based baselines is that traditional sequence classification models struggle to converge when decoding sentiment directly from EEG features. As shown in scores from \textit{Bi-LSTM} and \textit{MLP}, they hardly outperform random guess. And we can see that the \textit{Bi-LSTM} model can not take advantage of the additional sequential information from EEG feature sequences, since we do not observe improvements compared with simply doing classification on a single averaged feature vector using \textit{MLP}. By comparing scores from \textit{BERT$_{rand}$} and \textit{BERT$_{fine}$}, we find that directly applying pre-trained language model to EEG features does not help much. We hypothesize that since the amount of labeled data is very limited ($\sim$ 400 unique sentences) compared with a typical sentiment dataset such as Stanford Sentiment Treebank~\cite{socher2013recursive}($\sim$ 10k unique sentences), it is simply too noisy for the models to extract useful features out of the EEG sequences. We qualitatively verify this hypothesis by observing high bias in the predicted label distribution, as shown in Table~\ref{tab:label_distribution}.
\input{pred_label_distribution}

On the other hand, we find that our best-performed zero-shot pipeline (\textit{DEC$_{BART}$+CLS$_{BART}$}) significantly outperforms all fully supervised baselines by a large margin of over 20$\%$. We also observe that our pipeline is more robust to noisy data as shown in Table~\ref{tab:label_distribution}. As discussed in Section \ref{sec:method_zeroshot}, we think the reason why our zero-shot pipeline works much better than baselines is that, by first decoding EEG sequence into text, the decoder filters out noisy information from the EEG features and enables the model to effectively leverage the power of text-based classifiers. Furthermore, we show that our zero-shot pipeline is highly modularized in the sense that improving a single component can result in improvements on the whole model. For example, the BART-based classifier (\textit{CLS$_{BART}$}), which achieves top performance on text input, also constitutes our best-performed pipeline (\textit{DEC$_{BART}$+CLS$_{BART}$}). The results from the crippled versions of decoder (\textit{DEC$_{BART}$}) show that the pretrained language model and the additional Multi-layer TransformerEncoder (MTE) are essential to the pipeline, as we observe substantial performance drop on the ones \textit{w/o MTE}, and \textit{w/o pretrained weights}.

%% file: dataset.tex
\begin{table}[t]
\small
\centering
\begin{tabular}{p{1.5cm}p{1.5cm}p{1.5cm}p{1.5cm}}
\toprule
\textbf{Reading Task} & \textbf{\#Unique Sentences} & \textbf{\#Training Samples} & \textbf{\#Testing Samples} \\ \midrule
SR v1.0 & 400 & 3391 & 418\\
NR v1.0 & 300 & 2406 & 321\\
NR v2.0 & 349 & 4456 & 601\\
TSR v1.0 & 407 & 3372 & 350  \\
\bottomrule
\end{tabular}
\caption{\label{tab:dataset} Dataset Statistics. SR: Normal Reading (sentiment), NR: Normal Reading (wikipedia), TSR: Task Specific Reading (wikipedia). We used data from 12 subjects in v1.0 and 18 subjects in v2.0.}
\end{table}

%% file: generation_scores.tex
\begin{table*}[t]
\small
\centering
\begin{tabular}{l|c| c c c c | c c c}
\toprule
\multirow{1}{*}{\textbf{Reading}} &
\multirow{1}{*}{\textbf{\#Training}} &
\multicolumn{4}{c|}{\textbf{BLEU-N}($\%$)} &
\multicolumn{3}{c}{\textbf{ROUGE-1}($\%$)}
\\
\textbf{Task}
& \textbf{Sample} 
& \textbf{ $N=1$ } & \textbf{ $N=2$ } & \textbf{ $N=3$ } & \textbf{ $N=4$ }
& P & R & F
\\
\midrule
SR v1.0 (half$_{1st}$)
& 1771
& 33.5
& 16.6
& 7.2
& 3.4
& 22.8
& 21.6
& 22.0
\\
SR v1.0 (half$_{1st}$) + NR v1.0 (half$_{1st}$)
& 3129
& 37.3
& 20.2
& 10.2
& 5.0
& 27.8
& 25.4
& 26.5
\\
SR v1.0 (half$_{1st}$) + NR v1.0 (half$_{2nd}$)
& 3058
& 37.4
& 18.8
& 8.9
& 4.0
& 26.3
& 25.2
& 25.5
\\
SR v1.0
& 3391
& 34.5
& 17.2
& 7.2
& 3.0
& 23.9
& 22.0
& 22.8
\\
SR v1.0 $+$ NR v1.0  
& 5797
& 37.1
& 20.0
& 10.4
& 5.3
& 27.7
& 26.0
& 26.8
\\
SR v1.0 $+$ NR v1.0 $+$ TSR v1.0  
& 9169
& 37.4
& 20.0
& 10.5
& 5.7
& 27.4
& 24.6
& 25.9
\\
SR v1.0 $+$ NR v1.0 $+$ NR v2.0  
& 10710
& \textbf{40.1}
& \textbf{23.1}
& \textbf{12.5}
& \textbf{6.8}
& \textbf{31.7}
& \textbf{28.8}
& \textbf{30.1}
\\
\hspace{1em} w/o pretrained weights
& 10710
& 24.7
& 7.3
& 2.4
& 1.0
& 19.4
& 20.2
& 18.9
\\


\bottomrule
\end{tabular}
\caption{\label{tab:generation_scores} EEG-To-Text sequence-to-sequence model evaluation: (half$_{1st}$) and (half$_{2nd}$) indicate using data from the first half of or the second half of the subjects respectively; no parenthesis means using data from all subjects.}
\end{table*}


%% file: generation_results.tex
\begin{table*}[t]
    \small
    \centering
    \begin{tabular}{m{1em}|m{48em}}
    \toprule 
     \multirow{2}{*}{(1)} 
     & Ground Truth: He is a prominent \textbf{member of} the \textit{Bush family}, the younger brother of \textbf{President George W. Bush}...\\
     \cmidrule{2-2}
     & Model Output: was a former \textbf{member of} the \textit{American family}, and son brother of \textbf{President George W. Bush}...\\
     \midrule
     
     \multirow{2}{*}{(2)} 
     & Ground Truth: \underline{Raymond Arrieta} (born March 26, 1965 in \underline{San Juan}, \textbf{Puerto Rico}) is considered by many to be one of \textbf{Puerto Rico's greatest} \underline{comedians}.\\
     \cmidrule{2-2}
     & Model Output: \underline{mond wasaga},19 in 17, 18) \underline{New Francisco}, \textbf{Puerto Rico}) is a one many to be the of the \textbf{Rico's greatest} \underline{poets}.\\
     \midrule
     
     \multirow{2}{*}{(3)} 
     & Ground Truth: He was first \textit{appointed} to fill the Senate \textbf{seat} of \underline{Ernest Lundeen} who had \textbf{died} in office.\\
     \cmidrule{2-2}
     & Model Output: was a \textit{elected} to the the position \textbf{seat} in the \underline{Hemy} in \textbf{died} died in 18 in\\
     \midrule
     
     \multirow{2}{*}{(4)} 
     & Ground Truth: \underline{Adolf Otto Reinhold Windaus} (December 25, 1876 - June 9, 1959) was a significant \textit{German chemist}.\\
     \cmidrule{2-2}
     & Model Output: rian \underline{Hitler},hardt,eren18 18, 1885 – January 3, 18) was a \textit{German figure}- and
     \\
     \midrule
     
     \multirow{2}{*}{(5)} 
     & Ground Truth: It's \textit{not a particularly} \textbf{good} film, but neither is it a \textit{monsterous }one.\\
     \cmidrule{2-2}
     & Model Output: was a a \textit{bad} \textbf{good} story, but it is it \textit{bad bad}. one.\\
     
    
    \bottomrule
    \end{tabular}
    \caption{EEG-To-Text decoding examples on unseen test sentences. (1-4) are biographical sentences from Wikipedia in NR v1.0, v2.0. (5) is movie review in SR v1.0. \textbf{Bold} words indicate exact match, \textit{Italic} words indicate semantic resemblance, and \underline{Underline} words indicate error match.}
    \label{tab:generation_results}
\end{table*}


%% file: entity_matching.tex
\begin{table}[ht]
\small
\centering
\begin{tabular}{l|c|c}
\toprule
\textbf{Reading Task} & \textbf{LCS($\%$)} & \textbf{Multiset($\%$)}\\ \midrule
SR v1.0 & 14.9 & 17.8\\
SR v1.0 $+$ NR v1.0 & 29.2 & 50.2 \\
SR v1.0 $+$ NR v1.0 $+$ NR v2.0 & \textbf{35.6} & \textbf{55.7} \\
\bottomrule
\end{tabular}
\caption{\label{tab:entity_matching} Named entity type matching results. \textit{LCS} refers to Longest Common Subsequence based matching score, \textit{Multiset} refers to Multiset Cosine Similarity.
}
\end{table}

%% file: sentiment_scores.tex
\begin{table*}[t]
\small
\centering
\setlength\tabcolsep{6pt}
\setlength\extrarowheight{2pt}
\begin{tabular}{l|c|c|c c c c}
\toprule
\multirow{2}{*}{\textbf{Model}}
& \textbf{Test} 
& \textbf{Is}
& \multirow{2}{*}{\textbf{Precision($\%$)}} 
& \multirow{2}{*}{\textbf{Recall($\%$)}}
& \multirow{2}{*}{\textbf{F1($\%$)}}
& \multirow{2}{*}{\textbf{Accuracy($\%$)}}
\\
\multirow{1}{*}{}
& \textbf{Input} 
& \textbf{Zero-Shot}
& \multirow{1}{*}{}
& \multirow{1}{*}{}
& \multirow{1}{*}{}
& \multirow{1}{*}{}
\\
\midrule
MLP~\cite{hollenstein2019cognival}
& $\hat{\textbf{e}}$
& No
& 32.8
& 33.6
& 27.5
& 31.8
\\
Bi-LSTM~\cite{hollenstein2021decoding}
& $\mathcal{E}$
& No
& 33.9
& 34.1
& 17.4
& 30.9
\\
BERT$_{rand}$
& $\mathcal{E}$
& No
& 38.2
& 33.5
& 30.0
& 35.5
\\
BERT$_{fine}$
& $\mathcal{E}$
& No
& 23.7
& 34.5
& 27.2
& 36.6
\\
\midrule
DEC$_{BART}$ $+$ CLS$_{BERT}$
& $\mathcal{E}$
& Yes
& 61.0
& 50.4
& 50.1
& 49.1
\\
DEC$_{BART}$ $+$ CLS$_{RoBERTa}$
& $\mathcal{E}$
& Yes
& 58.2
& 51.2
& 51.9
& 50.9
\\
DEC$_{BART}$ $+$ CLS$_{BART}$
& $\mathcal{E}$
& Yes
& \textbf{62.4}
& \textbf{56.5}
& \textbf{55.6}
& \textbf{55.3}
\\
\hspace{1em} w/o pretrained $+$ CLS$_{BART}$
& $\mathcal{E}$
& Yes
& 10.0
& 33.3
& 15.4
& 30.0
\\
\hspace{1em} w/o MTE $+$ CLS$_{BART}$
& $\mathcal{E}$
& Yes
& 41.3
& 40.9
& 39.3
& 40.4
\\
\midrule
\midrule
CLS$_{BERT}$
& $\mathcal{S}$
& No
& 76.0
& 74.5
& 74.1
& 75.4
\\
CLS$_{RoBERTa}$
& $\mathcal{S}$
& No
& 72.5
& 71.3
& 70.6
& 72.7
\\
CLS$_{BART}$
& $\mathcal{S}$
& No
& \textbf{79.3}
& \textbf{78.3}
& \textbf{77.4}
& \textbf{79.7}
\\

\bottomrule
\end{tabular}
\caption{\label{tab:sentiment_scores} Ternary sentiment classification results on SR v1.0 testset. In \textit{Test Input} column, $\hat{\textbf{e}}$ means aggregated sentence-level EEG features, $\mathcal{E}$ is word-level EEG feature sequence, $\mathcal{S}$ is text sentences. In the Model column, the first section contains baselines explicitly trained on EEG-Sentiment pairs, where subscript $_{rand}$ indicates it's randomly initialized, and $_{fine}$ indicates it's fine-tuned from pretrained checkpoint. The second section contains our proposed Zero-shot pipelines with different choices of decoder (DEC) and classifier (CLS). The third section contains text-based baselines.}
\end{table*}

%% file: pred_label_distribution.tex
\begin{table}[!ht]
\small
\centering
\begin{tabular}{l|c c c}
\toprule
\textbf{Model} & \textbf{Pos($\%$)} & \textbf{Neu($\%$)} &\textbf{Neg($\%$)}\\ \midrule
BERT$_{rand}$ & 68.6 & 26.5 & 4.9\\
BERT$_{fine}$ & 72.4 & 27.0 & 0.6\\
\midrule
DEC$_{BART}$+CLS$_{BART}$ & 19.3 & 54.4 & 26.3\\
\midrule
Ground Truth & 37.5 & 30.0 & 32.5\\
\bottomrule
\end{tabular}
\caption{\label{tab:label_distribution} Predicted label distribution: \textit{Pos}, \textit{Neu}, \textit{Neg} means \textit{Positive}, \textit{Neutral}, \textit{Negative} respectively. Model names are consistent with Table~\ref{tab:sentiment_scores}.}
\end{table}

%% file: 4related_works.tex
\section{Related Work}
\label{sec:related_works}
Related work on brain-to-speech and brain-to-text decoding can be categorized into three major approaches by the features they are capturing: motor imagery based,
overt speech based,
and inner speech based.
Various kinds of brain signals have been explored including EEG, ECoG, and fMRI. We point out that previous approaches still have limitations in terms of vocabulary size, articulation dependency, speed and device. Additional details can be found in Appendix.

Motor imagery based systems, e.g., point-and-click~\cite{pandarinath2017high,jarosiewicz2015virtual} and imaginary handwriting~\cite{willett2021high} has high accuracy but relatively low typing rate. Overt speech based approaches for decoding or synthesizing speech achieve faster communication rate. However, these approaches either require subjects to physically move their vocal tract~\cite{herff2015brain, anumanchipalli2019speech, makin2020machine}, or require the subjects to imagine the physical articulation of the sentence~\cite{doi:10.1056/NEJMoa2027540}. This makes the decoding system language-dependent, since the same concept may have totally different pronunciations in different languages. Another line of work tries to address articulation dependency by decoding language from imagined speech or read text~\cite{panachakel2021decoding, sun2019towards, nieto2021thinking}. A major limitation for most of the aforementioned approaches is that current experiments are often constrained with a small closed vocabulary, with ten to a few hundred unique words. Moreover, most state-of-the-art high-performance brain-computer-interface~\cite{willett2021high, makin2020machine,pandarinath2017high} for language communication use invasive devices such as Electrocorticography (ECoG). Although compared to non-invasive devices like Electroencephalogram (EEG), ECoG has a higher temporal and spatial resolution as well as a higher signal-to-noise ratio, it is hard to collect large-scale dataset, and is currently impractical to extend the benefit to healthy people.

Our work is distinguished from previous work by the fact that we decode text from EEG features in a Sequence-To-Sequence manner over the whole English vocabulary. And our model is able to decode unseen sentences and handle EEG data from various sources and subjects. 

Related work on sentiment/emotion analysis from brain signals traditionally focuses on video or image stimulus~\cite{koelstra2011deap, liu2020eeg}. Previous attempts using text-elicited brain signals often treat brain signals only as an additional input along with other traditional modalities like text and image~\cite{hollenstein2021decoding, kumar2019fusion, gauba2017prediction}. 
 
Our work on zero-shot sentiment discovery extends previous work by using only EEG features as input and by the fact that we do not require any EEG-Sentiment labeled pairs. 

%% file: 5conclusion.tex
\section{Conclusions and Future Work}
In this paper, we introduce two challenging tasks, namely, open vocabulary EEG-To-Text Sequence-To-Sequence decoding and EEG-based sentence sentiment classification. We propose novel frameworks leveraging pretrained language models which show great scalability and zero-shot ability. Future work is needed on collecting larger-scale EEG-Text datasets as well as extending the current framework to multilingual settings. Furthermore, previous studies~\cite{gernsbacher2003neuroimaging, segaert2012shared} demonstrated that certain frontal regions, e.g., Broca's area, show activation during both language comprehension and language production processes. One future direction is to apply our current model on decoding inner speech in an open vocabulary setting. A dedicated sentence-level inner speech dataset with a larger vocabulary size is needed, since current datasets~\cite{nieto2021thinking} on inner speech decoding have very limited word coverage. 

%% file: Appendix_section.tex
\section{Implementation Details}
For our EEG-To-Text Sequence-To-Sequence model, the input embedding dimension is 840 ($8\times105$). The hidden dimension of the pretrained BART\footnote{\url{https://huggingface.co/facebook/bart-large}} is 1024. The additional multi-layer TransformerEncoder has 6 layers, each with 8 heads and a hidden dimension of 2048. Instead of using the two-step training approach described in~\cite{lewis2019bart}, we use one-step training with a learning rate of 5e-7. We find that in our setting, one-step training achieves equally good results with less training time.
For zero-shot EEG-based sentiment classification, we train text-based classifiers by finetuning pretrained BERT\footnote{\url{https://huggingface.co/bert-base-cased}}~\cite{devlin2018bert}, RoBERTa\footnote{\url{https://huggingface.co/roberta-base}}~\cite{liu2019roberta}, BART\footnote{\url{https://huggingface.co/facebook/bart-large}}~\cite{lewis2019bart} on external StanfordSentimentTreebank~\cite{socher2013recursive} dataset with learning rates as 1e-3, 1e-3, 1e-4 respectively. The EEG-based supervised baselines (MLP, Bi-LSTM and BERT) are trained on SR v1.0 EEG-Sentiment pairs with learning rates as 1e-3, 1e-3, 5e-5. All models are trained on one Tesla V100 GPU with 16GB DRAM. All training trains for 25 epoches with a batch size of 32. We use SGD\footnote{\url{https://pytorch.org/docs/stable/generated/torch.optim.SGD.html}} optimizer and StepLR\footnote{\url{https://pytorch.org/docs/stable/generated/torch.optim.lr_scheduler.StepLR.html}} scheduler with a step size of 20. The best model is selected based on performance on validation set. All random seeds are set to 312.
\section{Details About EEG Input Data}
In ZuCo dataset~\cite{hollenstein2018zuco, hollenstein2020zuco}, EEG features are extracted into 8 frequency bands by doing band-pass filtering and Hilbert transform on preprocessed raw EEG signals.  Each word-level EEG feature has a fixed dimension of 105. For each token in an EEG sequence, we concatenate its feature from all 8 frequency bands, i.e., theta1 (4–6Hz), theta2 (6.5–8 Hz), alpha1 (8.5–10 Hz), alpha2 (10.5–13 Hz), beta1 (13.5–18Hz), beta2 (18.5–30 Hz), gamma1 (30.5–40 Hz) and gamma2 (40–49.5 Hz), into one feature vector with a dimension of 840. Word-level EEG features are obtained by synchronizing and aggregating raw EEG features with fixations on each word. When the subject is reading a sentence, the number of fixations on each word is recorded by eye-tracking device with a predefined duration threshold.
ZuCo dataset provides several types of EEG features corresponding to different extracted eye-tracking features, including first fixation duration (FFD), total reading time (TRT), gaze duration (GD), single first fixation (SFD) and go-past time (GPT). We use EEG features aggregated on gaze duration (GD) in our experiment.
Note that there can be no fixation on a certain word in a sentence, thus, the length of a text sequence can be different from the length of the corresponding EEG feature sequence. 
\section{Details About Related Work}
A typical speaking rate for an English speaker is 4 syllables per second~\cite{cruttenden2014gimson}, which results in 100-130 words per minute. Current point-and-click communication interfaces ~\cite{pandarinath2017high,jarosiewicz2015virtual} has a typing rate of 10-30 characters per minute. A recently proposed approach for brain-to-text communication by imagining handwritten letters~\cite{willett2021high} increases the typing rate to 115 characters per minute, but still leaves a big gap towards natural communication. 

In overt speech based approaches that require subjects to physically or imaginarily move their vocal tract~\cite{herff2015brain, anumanchipalli2019speech, makin2020machine, doi:10.1056/NEJMoa2027540}, the model captures low-level auditory features such as single phones elicited from the movements of the one's tongue, lips, jaw, etc. Thus, these approaches depend on particular languages and potentially accents.

Currently, in most brain-to-speech and brain-to-text experiments, training and testing sentences can come from the same set of unique sentences~\cite{makin2020machine}. In terms of handling unseen sentences or words, only pairwise matching~\cite{sun2019towards} has been evaluated, no Sequence-To-Sequence generation on unseen sentences has been explored. 